%% file: main.tex
\documentclass{article}
 
\newcommand\ie{\emph{i.e.}}

\newcommand\etc{\emph{etc.}}
 
\newcommand\etal{\emph{et al.}}
\newcommand{\Rows}[1]{\multirow{2}{*}{#1}}
\newlength\savewidth

\makeatletter\renewcommand\paragraph{\@startsection{paragraph}{4}{\z@}
  {.5em \@plus1ex \@minus.2ex}{-.5em}{\normalfont\normalsize\bfseries}}\makeatother

\usepackage{microtype}
\usepackage{subfigure}
\usepackage{amsfonts}
\usepackage{booktabs} 
\usepackage{graphicx}
\usepackage{color}
\usepackage{multicol, blindtext}
\usepackage{booktabs}
\usepackage{soul}
\usepackage{csquotes}
\usepackage{multirow}
\usepackage{amsmath}
\usepackage{amssymb}
\usepackage{tabulary}
\usepackage[dvipsnames]{xcolor}
\usepackage{cancel}
\definecolor{deemph}{gray}{0.6}
\newcommand{\gc}[1]{\textcolor{deemph}{#1}}
\definecolor{green}{HTML}{39b54a}  
\definecolor{greenn}{HTML}{008000}  
\definecolor{blue}{HTML}{1F51FF}  %

\newcommand{\bl}[1]{\textcolor{blue}{#1}}
\newcommand{\re}[1]{\textcolor{red}{#1}}
\makeatother

\def\Vec#1{{\boldsymbol{#1}}}

\usepackage{hyperref}

\usepackage[accepted]{icml2022}

\icmltitlerunning{You Only Cut Once: Boosting Data Augmentation with a Single Cut}

\begin{document}

\twocolumn[
\icmltitle{You Only Cut Once: Boosting Data Augmentation with a Single Cut }




\begin{icmlauthorlist}
\icmlauthor{Junlin Han}{d61,anu,ade}
\icmlauthor{Pengfei Fang}{d61,anu}
\icmlauthor{Weihao Li}{d61}
\icmlauthor{Jie Hong}{d61,anu}
\icmlauthor{Mohammad Ali Armin}{d61}
\\
\icmlauthor{Ian Reid}{ade}
\icmlauthor{Lars Petersson}{d61}
\icmlauthor{Hongdong Li}{anu}
\end{icmlauthorlist}

\icmlaffiliation{d61}{Data61-CSIRO, Canberra, Auistralia}
\icmlaffiliation{anu}{Australian National University, Canberra, Australia}
\icmlaffiliation{ade}{University of Adelaide, Adelaide, Australia}

\icmlcorrespondingauthor{Junlin Han}{junlin.han@data61.csiro.au}

\icmlkeywords{Machine Learning, ICML}

\vskip 0.3in
]



\printAffiliationsAndNotice{}  

\begin{abstract}
We present \textbf{Y}ou \textbf{O}nly \textbf{C}ut \textbf{O}nce (YOCO) for performing data augmentations. YOCO cuts one image into two pieces and performs data augmentations individually within each piece. Applying YOCO improves the diversity of the augmentation per sample and encourages neural networks to recognize objects from partial information. YOCO enjoys the properties of parameter-free, easy usage, and \textbf{boosting almost all augmentations for free}. Thorough experiments are conducted to evaluate its effectiveness. We first demonstrate that YOCO can be seamlessly applied to varying data augmentations, neural network architectures, and brings performance gains on CIFAR and ImageNet classification tasks, sometimes surpassing conventional image-level augmentation by large margins. Moreover, we show YOCO benefits contrastive pre-training toward a more powerful representation that can be better transferred to multiple downstream tasks. Finally, we study a number of variants of YOCO and empirically analyze the performance for respective settings. 
Code is available at {\href{https://github.com/JunlinHan/YOCO}{\re{GitHub}}}.

\end{abstract}
\section{Introduction}

\begin{figure}[htb]
     \centering
     \includegraphics[width = 8cm]
     {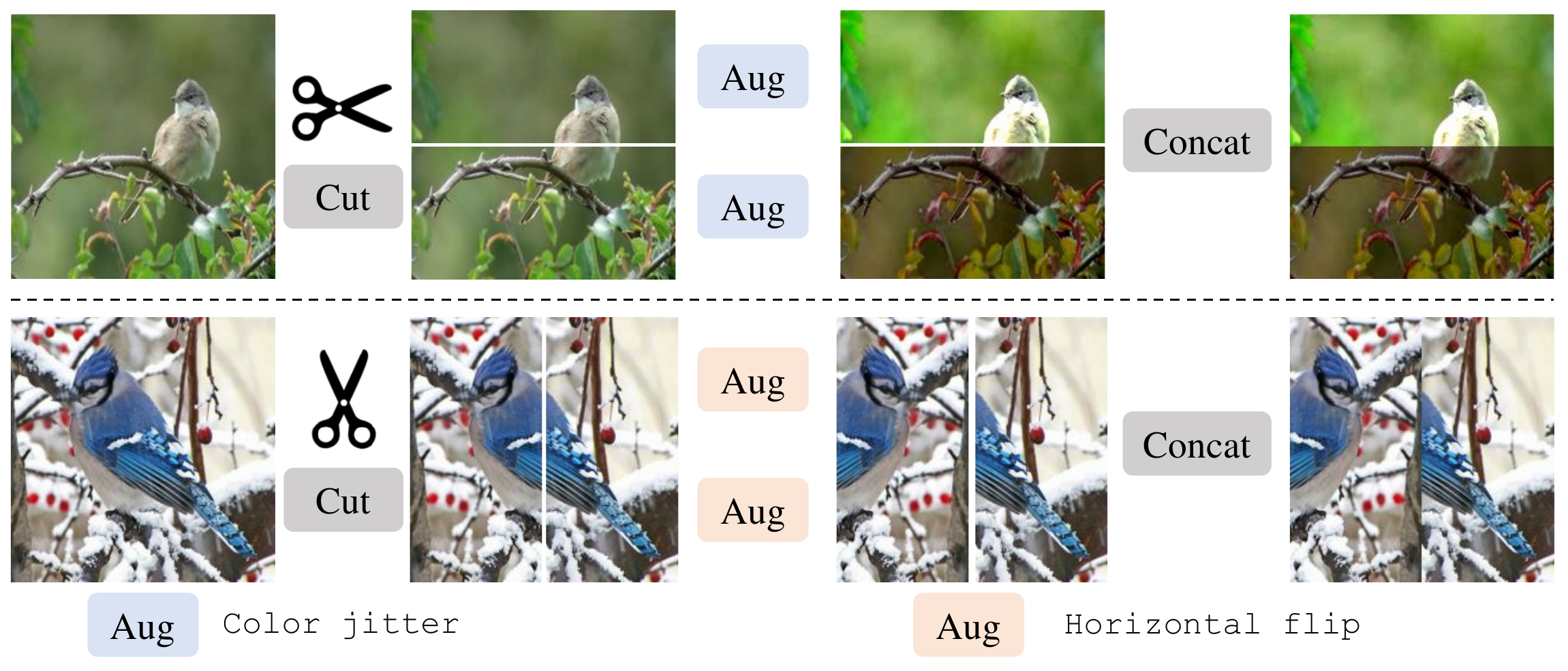}
     \caption{YOCO cuts one image into two equal pieces, either in the height or the width dimension. The same data augmentations are performed independently within each piece. Augmented pieces are then concatenated together to form one single augmented image. The upper row shows the results of YOCO applied to \texttt{Color jitter}, where the two pieces are both augmented, producing a diversified fully-augmented image. The lower row presents the result of employing YOCO to \texttt{Horizontal flip}, only the left piece has been augmented, yielding a partially-augmented image.   }
     \label{fig:ppline}
\end{figure}

Deep neural networks have been widely applied to various computer vision tasks such as image classification~\cite{resnet,alexnet,vgg}, semantic segmentation~\cite{FCN,deeplab}, and object detection~\cite{he2017mask,RCNN}. However, with the rapid gains in computational resources, neural networks can easily overfit a training set with numerous images~\cite{imagenet}, leading to a large generalization gap on test data. In addition, neural networks can be effortlessly fooled by adversarial examples or attacks~\cite{hendrycks2021natural, pgd}. Thus, to improve the generalization performance and robustness of neural networks, multiple training strategies have been proposed mainly from two views: regularization techniques~\cite{srivastava2014dropout, huang2016deep,szegedy2016rethinking} and data augmentations~\cite{alexnet}. In this work, we focus on improving data augmentations in particular.

\begin{figure*}[!htb]
     \centering
     \includegraphics[width = 17cm]
     {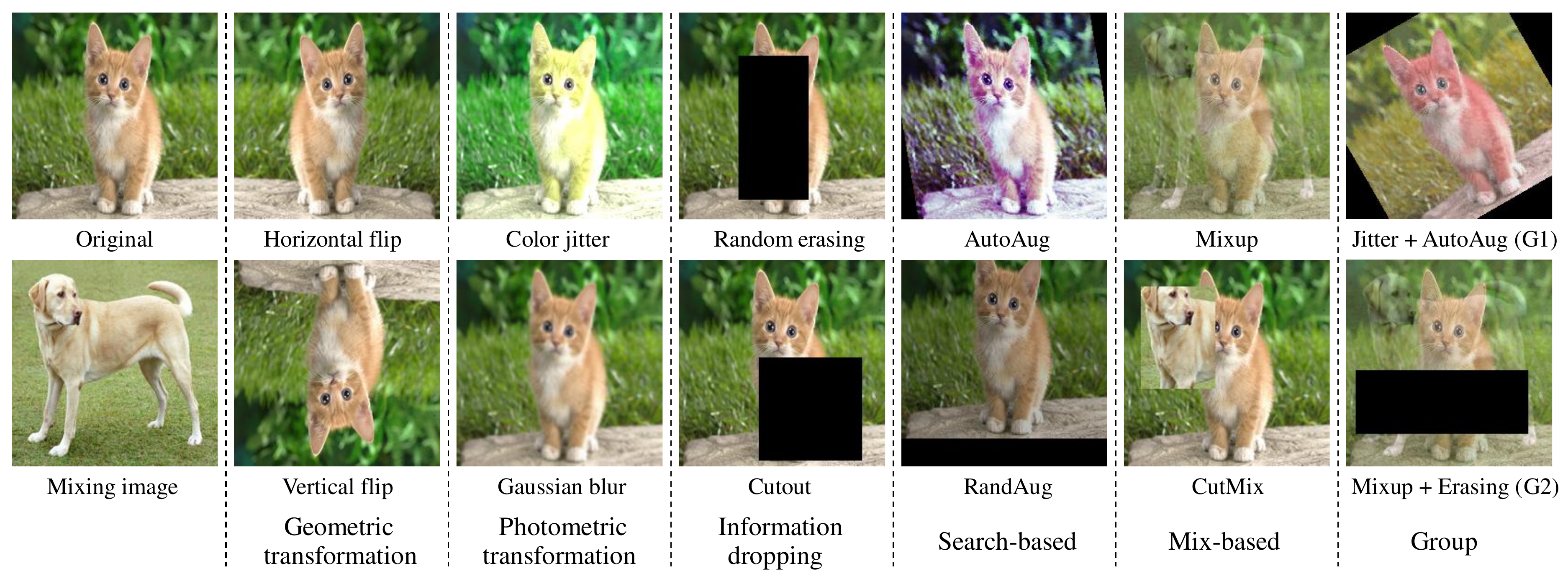}
     \caption{Illustrations of the studied augmentations in classification tasks. We roughly classify data augmentations into five categories. For each category, we study the two most representative augmentations. We also combine two augmentations as a group, forming a new category named group of augmentations. }
     \label{fig:aug_overview}
\end{figure*}

Standard data augmentations are mostly performed at the image-level, which yields all-round performance improvements in both generality and robustness. Usually, image-level augmentation preserves semantics globally, following humans' cognitive intuition. Yet, human beings are also able to recognize objects from partial information alone. Patches, \ie{}, the internal information of images, are strong natural signals. Before the deep learning era, the power of patches has been exploited in many low-level vision~\cite{efros1999texture,kervrann2006optimal} and high-level vision~\cite{sivic2003video,csurka2004visual,lazebnik2006beyond} works. Recently, splitting one image into multiple non-overlapping patches as the input for neural networks is a key success of Vision Transformer (ViT)~\cite{dosovitskiy2020image}. 
However, how to perform data augmentations at a non-image level (in other words, patch-level or piece-level) is rarely studied.

From the above, we hypothesize that it is desirable to propose a strategy of performing augmentations beyond the image level. An image can be seen as a combination of multiple patches, or even two pieces. For a specific augmentation, we may perform this augmentation on these pieces individually and combine transformed pieces back to a single image. Such a strategy should increase the diversity in both local regions level as well as at the holistic image level and may also encourage neural networks to share the same cognitive ability of recognizing objects from partial information like humans can.

Following our hypothesis, we propose You Only Cut Once (YOCO), a simple method of performing data augmentations. Specifically, YOCO cuts one image into two equal pieces, in either the height dimension or the width dimension, performs data augmentations individually within each piece, and concatenates the two augmented pieces back together, as shown in Figure~\ref{fig:ppline}.

We conduct extensive evaluations of YOCO on various augmentations, neural network architectures, datasets, and challenging tasks. For image classification tasks on CIFAR-10 and CIFAR-100, YOCO can be seamlessly applied to almost all (\textbf{11} out of \textbf{12}) data augmentations, where we evaluate YOCO with \textbf{7} neural network architectures and \textbf{12} augmentations. YOCO boosts\textbf{ 11 }augmentations to train classification models with better generalization abilities. 
For ImageNet, we evaluate generality, partial image recognition, calibration, robustness against adversarial attacks, corruption robustness, and robustness under distribution shifts. The six most representative augmentations are evaluated (the first row of Figure~\ref{fig:aug_overview}). For every evaluation metric, YOCO outperforms image-level augmentation overall.

Next, we study the effect of YOCO on contrastive pre-training
and conduct experiments based on MoCo v2~\cite{chen2020improved} and SimSiam~\cite{chen2021exploring}. 
YOCO benefits the contrastive self-supervised representation learning, where a more powerful representation is produced, which can be better transferred to classification, object detection, and instance segmentation tasks. 

YOCO is parameter-free, easy to employ, and \textbf{boosts almost all augmentations for free}. As the presented YOCO is with the most simplified setting, we study some more complex design choices with more hyper-parameters embedded. Lastly, we analyze and study four questions (1): how does YOCO work? (2): when to employ YOCO? (3): why should we employ YOCO? and (4): does YOCO help other vision tasks?

\section{Related work}

\subsection{Data Augmentation}
Data augmentation is an effective strategy for training neural networks. We roughly divide common data augmentations into five categories, and for each category, we choose the two most representative augmentations: (1) Geometric transformation (\texttt{Horizontal flip} and \texttt{Vertical flip}), (2) Photometric transformation (\texttt{Color jitter} and \texttt{Gaussian blur}), (3) Information dropping (\texttt{Random erasing}~\cite{zhong2020random} and \texttt{Cutout}~\cite{devries2017improved}), (4) Search-based (\texttt{AutoAug}~\cite{cubuk2018autoaugment} and \texttt{RandAug}~\cite{cubuk2020randaugment}), and (5) Mix-based (\texttt{Mixup}~\cite{zhang2017mixup} and \texttt{CutMix}~\cite{yun2019CutMix}). 
To study the effect of YOCO applied to multiple augmentations, we further combine two augmentations as a group, forming category (6) group of augmentations (\texttt{Color jitter + AutoAug} (\texttt{G1}) and \texttt{Mixup + Random erasing} (\texttt{G2})). 
Figure~\ref{fig:aug_overview} presents all augmentations we studied. We leave feature-level augmentation~\cite{upchurch2017deep, wang2021regularizing} and GAN-based augmentation~\cite{bowles2018gan} for future study.

Search-based augmentation utilizes reinforcement learning to search from a pool of augmentation policies for an optimal combination. Mix-based augmentation employs multi-image information by creating mixed input images with soft labels for training. Augmentations are important for multiple vision tasks. For instance, recent contrastive learning methods~\cite{he2020momentum,chen2020simple,chen2020improved,chen2021exploring} relies heavily on multiple data augmentations to construct contrastive pairs.
All these augmentations are previously performed at the image-level, we study how to perform augmentations at a non-image level in multiple vision tasks.

\subsection{Patches in Vision}
Patches have been widely used as strong signals for various vision tasks, in both conventional methods, and learning-based methods. Applications include texture synthesis~\cite{efros1999texture}, image denoising~\cite{kervrann2006optimal}, image-to-image translation~\cite{park2020contrastive,han2021dcl}, super-resolution~\cite{shocher2018zero}, bag-of-features based classification~\cite{sivic2003video,csurka2004visual,lazebnik2006beyond}. Recently, both CNNs and ViTs leverage patches as the input for classification networks~\cite{brendel2019approximating,dosovitskiy2020image}. 

Patches are also employed in data augmentation. For instance, \texttt{Patch Gaussian}~\cite{lopes2019improving} applies \texttt{Gaussian Blur} to only a portion of the images. 
Cut, Paste and Learn~\cite{dwibedi2017cut} generates large annotated instance datasets by cutting object instances and pasting them on backgrounds. Georgakis \etal{}~\cite{georgakis2017synthesizing} superimposes 2D object images into real environments images.
\texttt{CutMix}~\cite{yun2019CutMix} replaces one patch of an image with a patch from another image.
PAA~\cite{lin2021local} extends the setting of \texttt{AutoAug}~\cite{cubuk2018autoaugment} by searching augmentation policies in pre-defined patches. Qin \etal{}~\cite{qin2021understanding} improves the robustness of ViTs through patch-based negative augmentation. However, no existing work has studied how to generally perform the same augmentation at a non-image level.

\section{Method}
This section presents the proposed YOCO method.
Let $\Vec{X} \in \mathbb{R}^{C\times H \times W}$ denote an image and $a(\cdot)$ denote data augmentations, where $a(\cdot): \mathbb{R}^{C\times H \times W} \to \mathbb{R}^{C\times H \times W}, \Vec{X}' = a(\Vec{X})$. Here $\Vec{X}'$ is the augmented image and $a(\cdot)$ can be one of any data augmentations or multiple data augmentations.
In contrast to standard image-level augmentation, which applies the augmentation to the image directly {as $\Vec{X}' = a(\Vec{X})$}, YOCO first cuts the image into two equally sized pieces, in either height or width dimension with equal probability, as
\begin{equation*}
\resizebox{1.00\hsize}{!}{
$[\Vec{X}_1, \Vec{X}_2] = \mathrm{cut}_{\mathrm{H}}(\Vec{X}),\mbox{if $0 <p \leq 0.5$}, ~\text{s.t.}~\Vec{X}_i \in \mathbb{R}^{C \times \frac{H}{2} \times W}$
}
\end{equation*}
or
\begin{equation*}
\resizebox{1.00\hsize}{!}{
$[\Vec{X}_1, \Vec{X}_2] = \mathrm{cut}_{\mathrm{W}}(\Vec{X}),\mbox{if $0.5 <p < 1$}, ~\text{s.t.}~\Vec{X}_i \in \mathbb{R}^{C \times H \times \frac{W}{2}},$
}
\end{equation*}
where $p$ is sampled from $(0,1)$ uniformly (\ie, $p \in U(0, 1)$), $\Vec{X}_i$ are cut pieces, $\mathrm{cut}_{\mathrm{H}}(\cdot)$ and $\mathrm{cut}_{\mathrm{W}}(\cdot)$ represent the cut operation in the height dimension and the width dimension, respectively.

Then $a(\cdot)$ is applied separately within each piece, and augmented pieces are concatenated back together as $$\Vec{X}' = \mathrm{concat}[a_1(\Vec{X}_1), a_2(\Vec{X}_2)].$$

Augmentations are determined by randomness, which include random probabilities of being applied, random operations, and random magnitudes. Both $a_1(\cdot)$ and $a_2(\cdot)$ are instances of $a(\cdot)$, though applying the same data augmentation $a(\cdot)$, they may behave differently, thereby increasing the diversity in both the local region level and the holistic image level.

YOCO presented so far is in the most simple format, where the number of cuts and the position of the cut are both fixed. 
We also formulate a more general format of YOCO. In the general setting, You Only Cut Once (YOCO) becomes You Cut Many Times, where
$M$ cuts in the height dimension and $N$ cuts in the width dimension are applied, such that $(M+1) * (N+1)$ pieces of internal regions can be obtained as $[\Vec{X}_1, \Vec{X}_2, ..., \Vec{X}_{(N+1)*M}, \Vec{X}_{(N+1)*(M+1)}] = \mathrm{cut}^{M*N}(\Vec{X})$, where the position of cuts can be chosen uniformally or randomly. 
The identical procedure is also applied to those pieces, by $\Vec{X}' = \mathrm{concat}_{i =1, j = 1}^{M+1,N+1}\big[ a_{i*j} (\Vec{X}_{i*j}) \big].$

\input{tables/cifar10} 
\input{tables/cifar100}

\section{Classification}
In this section, we evaluate YOCO for its capability to improve image classifiers in multiple classification settings. Firstly, we study the effect of YOCO on CIFAR-10 and CIFAR-100~\cite{krizhevsky2009learning} datasets among 7 different models (5 CNN and 2 ViT) and 12 augmentations. Next, we validate YOCO on ImageNet-1K (henceforth referred to as ImageNet)~\cite{imagenet}. Due to that ImageNet is a large-scale dataset containing more than 1.2M training images from 1K categories, we generally and thoroughly evaluate YOCO on ImageNet to explore its potential.

\subsection{CIFAR Experiments}
\textbf{Experimental setup.} We train 5 CNN architectures (PreResNet18~\cite{he2016identity}, Xception~\cite{chollet2017xception}, DenseNet121~\cite{huang2017densely}, ResNeXt50~\cite{Xie2017}, WRN-28-10~\cite{Zagoruyko2016WRN}) and 2 ViT models (ViT~\cite{dosovitskiy2020image} and Swin~\cite{liu2021swin}) under the same training recipe, which mostly follows the training setting used in \texttt{Co-Mixup}~\cite{kim2021co}. That is, models are trained for 300 epochs, with an initial learning rate of 0.1 decayed by a factor 0.1 at epochs 100 and 200. We employ the SGD optimizer with a momentum of 0.9 and a weight decay of 0.0001. We use a batch size of 100, and models are trained on 2 GPUs. Training images are randomly cropped to 32 $\times$ 32 resolution with zero padding followed by random \texttt{Horizontal flip}. Other augmentations are added based on this setup.

\input{tables/imagenet_robust}

\textbf{Results.}
We report the \textit{best} results based the mean of three independent trials.
Table~\ref{tab:cifar10} and Table~\ref{tab:cifar100} present the results on CIFAR-10 and CIFAR-100, respectively. On CIFAR-10, among \textbf{12} augmentations, YOCO boosts \textbf{11} of them, toward a more generalized image classifier. For all \textbf{84} results, YOCO improves\textbf{ 76 }of them, showing the superiority generalization of YOCO over image-level augmentation. The improvements are non-trivial (average improvement $\geq$\textbf{ 0.2\%}) for \texttt{Blur}, \texttt{Cutout}, \texttt{AutoAug}, \texttt{Mixup}, \texttt{G1}, and \texttt{G2}.
On CIFAR-100, YOCO also boosts \textbf{11} augmentations, with\textbf{ 68 }results improved. The improvements of applying YOCO are notable (average improvement $\geq$ \textbf{0.7\%}) for \texttt{Blur}, \texttt{Erasing}, \texttt{Cutout}, \texttt{AutoAug}, \texttt{Mixup}, and \texttt{G1}. 

For both CIFAR-10 and CIFAR-100, we also observe that applying YOCO to \texttt{Horizontal flip} degrades the performance. We analyse the reasons empirically in Section~\ref{sec:Analysis}. The computational cost and memory usage of YOCO are both \textbf{negligible}. Compared to image-level augmentation, YOCO runs at an identical speed. Thereby, YOCO boosts nearly all augmentations for free. 

\subsection{ImageNet Experiments}
\textbf{Experimental setup.} 
We follow a simple training recipe used in \texttt{CutMix}~\cite{yun2019CutMix}. In doing so, we train a standard ResNet50~\cite{resnet} for 300 epochs, with an initial learning rate of 0.1, decayed by a factor of 0.1 at epochs 75, 150, and 225. We employ the SGD optimizer with a momentum of 0.9 and models are trained with a weight decay of 0.0001. We employ a batch size of 256, and 4 GPUs are used for training. Training images are randomly cropped to 224 $\times$ 224 resolution followed by random \texttt{Horizontal flip} prior to any other augmentations. We evaluate and test 6 representative augmentations as shown in the first row of Figure~\ref{fig:aug_overview}.

\subsubsection{Evaluation Protocol}
For models trained on ImageNet, other than generalization, we also evaluate partial image recognition, calibration, robustness against adversarial attacks, corruption robustness, and robustness under distribution shifts.

\textbf{Partial image recognition.} 
One design goal of YOCO is to inspire neural networks to share the same ability of recognizing objects from partial information like human can. We thus design a non-overlapping 4 crop evaluation. Specifically, we load test images in 512 $\times$ 512 resolution, and images are center cropped to 448 $\times$ 448 resolution, following the most common crop ratio of 0.875. We employ two cuts, one in the height dimension and one in the width dimension, to cut 448 $\times$ 448 images into 4 non-overlapping 224 $\times$ 224 resolution pieces. Four cropped pieces are evaluated and we report the mean of the results. 

\textbf{Calibration.} We follow the evaluation protocol used in \texttt{PixMix}~\cite{hendrycks2021pixmix}. The calibration task aims to match the empirical frequency of correctness. The posteriors from a model should satisfy $$
\mathbb{P}\left(Y=\arg \max _{i} f(X)_{i} \mid \max _{i} f(X)_{i}=C\right)=C,
$$ where $f$ is an image classifier, $X, Y$ are random variables representing the data distribution. We use RMS calibration error~\cite{hendrycks2018deep} and adaptive binning~\cite{nguyen2015posterior}.

\textbf{Adversarial attack.} We employ two adversarial attacks, FGSM~\cite{fgsm} and PGD~\cite{pgd}, to verify the robustness of image classifiers. We borrow the implementation from Torchattacks~\cite{kim2020torchattacks}. For all attacks, we apply an $\ell_{\infty}$ budget of 8/255. PGD is with 4 steps of optimization. We do not use larger steps of optimization as the accuracy can decline to zero for non-adversarially trained image classifiers. 

\textbf{Corruption robustness.} Following \texttt{Co-Mixup}~\cite{kim2021co}, we evaluate image classifiers on two challenging corrupted test sets, to verify the generalization ability and robustness on unseen environments. The corrupted test sets are created with the following operations: (1) replace the background 
with a random image and (2) adding \texttt{Gaussian noise} to the background.

\textbf{Distribution shift.}We employ ImageNet-A~\cite{hendrycks2021natural}, one of the most challenging ImageNet test sets containing natural adversarial examples, to verify the performance of image classifiers against input data distribution shifts. 

\subsubsection{Results} We report both \textit{best} and \textit{last} results. All results are summarized in Table~\ref{tab:imagenet_robust}. Visualizations on classification results are available in the Appendix~\ref{sec:appendix_visualization}.

\textbf{Generality.}
Among 5 tested augmentations, YOCO outperforms image-level augmentation in terms of generalization.  Considering its negligible cost, such improvement brought from YOCO is huge. For example, applying YOCO to \texttt{Horizontal flip} and \texttt{AutoAug} improves the \textit{last} top-1 accuracy by \textbf{0.42\%} and \textbf{0.72\%} respectively. 

\textbf{Partial image recognition.} 
YOCO improves the ability of image classifiers in recognizing objects from partial information. YOCO applied to 5 augmentations can reach improved performance. For \texttt{Random erasing}, dissimilar to other augmentations, applying YOCO does not involve changing the structure/consistency of images, thereby the partial image recognition ability is not improved. 

\textbf{Calibration.}
In terms of calibration, YOCO improves 4 augmentations. For \texttt{Horizontal flip}, the RMS of \textit{best} result is increased from\textbf{ 6.42} to \textbf{7.87}, while the \textit{last} result conversely benefits from YOCO. Epoch \textbf{171} produces the \textit{best} result for image-level augmentation while the \textit{best }result after applying YOCO is at epoch \textbf{235}. The inconsistent results are might due to classification models does not suffer from overfitting and overconfidence in the middle stage of training. 
To conclude, YOCO helps image classifiers to have better-calibrated predictions most of the time.

\textbf{Adversarial attack.}
For robustness against adversarial attacks, augmentations that do not involve color transformations usually benefit from YOCO, where YOCO hurts the robustness when applied to \texttt{Color Jitter}, \texttt{AutoAug}, and \texttt{G1}, but is helpful for all other augmentations. YOCO applied to color transformations breaks the color consistency of images. We suppose this property is not good for defending adversarial attacks. YOCO still performs well overall, as applying YOCO can be effectual for other augmentations, where YOCO approximately \textbf{doubles} the results of \texttt{Mixup} under PGD attack. 

\textbf{Corruption robustness.}
YOCO consistently shows better performance when evaluated with challenging corrupted images. 5 augmentations are boosted by YOCO. The improvements are particularly considerable for \texttt{AutoAug}, where the \textit{best} top-1 accuracy is improved with the performance margins of \textbf{0.54\%} and \textbf{1.58\% }for Random replacement and Gaussian noise correspondingly. This suggests that when only partial information is available, YOCO with better ability in recognizing partial objects behaves more robustly. 

\textbf{Distribution shift.}
Testing with the challenging ImageNet-A test set, YOCO again outperforms image-level augmentations overall. Improving the results on ImageNet-A is difficult, but applying YOCO to augmentations including \texttt{Horizontal flip}, \texttt{Color jitter}, \texttt{AutoAug}, and \texttt{G1} realizes improved results, showing the superiority of YOCO.

\textbf{Reported results.}
We find that the ResNet50 implemented in timm~\cite{rw2019timm}  yields a better performance than Pytorch/Torchvision inbuilt ResNet50 and original ResNet50 implemented in~\cite{resnet}. For a fair comparison, all ImageNet experiments are based on the timm implementation. For image-level augmentation, we report our reproduced results. Our reproduced results are usually higher than previously reported results. For instance, the \textit{best} Top-1 accuracy for \texttt{Horizontal flip} is reported as \textbf{77.1\%} by us, while being previously (same training recipe) reported as \textbf{76.3\%}~\cite{yun2019CutMix}.

\section{Contrastive Learning}
\input{tables/contrastive_transfer}
One core step of contrastive learning is to create two views from one image by data augmentations. YOCO increases the diversity of data, applying YOCO to contrastive learning yields more challenging views. However, strong augmentations are usually not beneficial to contrastive learning~\cite{chen2020simple}. In this section, we verify whether YOCO benefits MoCo v2~\cite{chen2020improved} and SimSiam~\cite{chen2021exploring} in the pre-training stage. The learned embedding is evaluated via multiple downstream tasks. Specifically, we evaluate the classification performance on ImageNet, under both the linear protocol and semi-supervised fine-tuning settings. The downstream tasks include VOC~\cite{everingham2010pascal} detection, COCO~\cite{lin2014microsoft} detection and COCO instance segmentation. 

\subsection{Experimental Setup}
\textbf{Pre-training.} 
We first apply YOCO to \texttt{Horizontal flip} only. Following the training setting introduced in MoCo v2 and SimSiam, we pre-train them with YOCO for 200 and 100 epochs respectively. The models are pre-trained following the same settings as their own base methods. The backbone is ResNet50. More details of training settings are provided in the Appendix~\ref{sec:appendix_contrastive}. 

\textbf{Linear protocol.}
We freeze the learned embedding and train a linear classifier on top of it. The experimental setting are identical to the same settings as their own base methods. 

\textbf{Fine-tuning.}
We also fine-tune the learned feature under the semi-supervised setting on ImageNet with 1\% and 10\% label fractions, where labels are provided by SimCLR~\cite{chen2020simple}. The fine-tuning setting follows Jigsaw Clustering~\cite{chen2021jigsaw}. 

\textbf{VOC detection. }
We mostly follow the setting used in MoCo. Faster R-CNN~\cite{ren2015faster} with the C4-backbone is fine-tuned on VOC 2007 trainval + 2012 train and evaluated on the VOC 2007 test. Due to a GPU limit, all experiments are conducted using 4 GPUs, whereas 8 GPUs are used in MoCo. We halve the batch size from 16 to 8, double the max iterations from 24K to 48K, and scale the base learning rate from 0.02 to 0.01. All reported results are the average over 3 trials under this setting. The results of MoCo v2 and SimSiam baseline are reproduced by us using the corresponding official pre-trained ResNet50 models.

\textbf{COCO detection and instance segmentation.}
Following MoCo, the model is Mask R-CNN~\cite{he2017mask} with a C4 backbone. Similar to VOC, batch size and base learning rate are halved while the maximum iteration is doubled, from 90K to 180K. The results of baselines are also reproduced by us under this setting. 

\subsection{Results}
Table~\ref{tab:contrastive} depicts the results. Among multiple tasks, both MoCo v2 and SimSiam benefit from YOCO. 
\input{tables/contrastive_supp}

The effect of YOCO applied to other augmentations is also studied, where we gradually increase the number of augmentations that are being applied with YOCO, specifically, we study YOCO applied to (1): \texttt{Horizontal flip}, (2): \texttt{Horizontal flip} +  \texttt{Gaussian blur}, (3): \texttt{Horizontal flip} + \texttt{Gaussian blur} + \texttt{Grayscale}, and (4): \texttt{Horizontal flip} + \texttt{Gaussian blur} + \texttt{Grayscale} + \texttt{Color jitter}. The results are presented in Table~\ref{tab:contrastive_supp}, where YOCO applied to more augmentations hurts the learned representation, suggesting that crafting more challenging views in moderation is helpful for contrastive representation learning.

Detailed training settings are provided in the Appendix~\ref{sec:appendix_contrastive}.

\section{Ablation study}
The concept of YOCO is presented in a simple way, where YOCO only involves one cut and the position of that cut is fixed. The general format of YOCO may involve multiple cuts and random positions of cuts. It naturally gives rise to questions like: Will it be beneficial if the position of the cut varies? How many cuts are needed to boost data augmentations? In this section, we aim to explore the optimal setting of YOCO. 

Experiments are performed on the CIFAR-100 dataset. All reported results are the mean of PreResNet18 and DenseNet121. We study \textbf{5} augmentations (the first row of Figure~\ref{fig:aug_overview}, except for \texttt{G1}). 

\textbf{Position of the cut.}
We employ the beta distribution $Beta(\alpha, \alpha)$ to sample the position of the cut, where we set $\alpha$ as 0.2, 0.4, 0.6, 0.8, and 1.0. The results are presented in Table~\ref{tab:position}. Compared to the fixed position, randomly sampled positions can be better applied to \texttt{Horizontal flip} and \texttt{AutoAug} but hurt the performance of being applied to \texttt{Color jitter}, \texttt{Erasing}, and \texttt{Mixup}. As the position of the cut behaves differently with augmentations, it is hard to select the optimal position. Designing an augmentation-specific strategy to select the optimal position of the cut might be a potential way to further improve the performance of YOCO.

\input{tables/position}

\textbf{Number of cuts.}
For both cuts in the height dimension and the width dimension, we set the number of cuts to 1, 2, and 3.
We report the results of all combinations in Table~\ref{tab:cutnumber}. For all augmentations, increasing the number of cuts hurts the performance. More cuts further increase the diversity of data, yielding the data to be too complex to learn from. This is particularly obvious when augmentations involve geometric transforms, as the semantic consistency is often corrupted when a large number of cuts are applied. For large-scale datasets such as ImageNet, we also observe a similar trend. Thus, You Only Cut Once is sufficient. 

\input{tables/cutnumber}

\section{Analysis and Discussion}
\label{sec:Analysis}

\subsection{How Does YOCO Work?}
We classify training data into two main categories, namely original data and augmented data, where the augmented data involves fully-augmented data, partially-augmented data, and diversified fully-augmented data. 
In most cases, YOCO is capable of creating two additional kinds of augmented data than image-level augmentation, partially-augmented data, and diversified fully-augmented data, as shown in Figure~\ref{fig:ppline}. We study the effect of YOCO by exploring the performance of classifiers trained with each kind of data solely. Results are summarized in Table~\ref{tab:solely}. 
\input{tables/solely}

Trained with one kind of data solely, partially-augmented data and diversified fully-augmented data both outperform fully-augmented data. This might be the reason for the superior performance of YOCO, where these two kinds of data further increase the diversity of augmented data, such that neural networks can learn more meaningful knowledge.

\subsection{When to Employ YOCO?}
We focus on studying when to employ YOCO in classification tasks. From both CIFAR and ImageNet results, YOCO outperforms image-level augmentation among most augmentations. However, YOCO applied to \texttt{Horizontal flip} hurts the performance of image classifiers on the CIFAR dataset. CIFAR mostly contains iconic object images, that is, the object, as a foreground part, occupies to most of the image. We hypothesise that YOCO applied to \texttt{Horizontal flip} ruins the structure/consistency of iconic object images. For more complex images (either more objects or more background components) such as ImageNet, YOCO can be applied to \texttt{Horizontal flip}, as the structural information for the foreground object can be kept in more cases, while the image diversity regarding background/object location is improved. Also, the \texttt{random resized crop} pre-processing plays an important role, as it may change a non-iconic object image to a iconic object image and vice versa. 

\begin{figure}[!htb]
     \centering
    \begin{minipage}[t]{0.49\linewidth} 
    \centering 
\includegraphics[width=4.9cm]{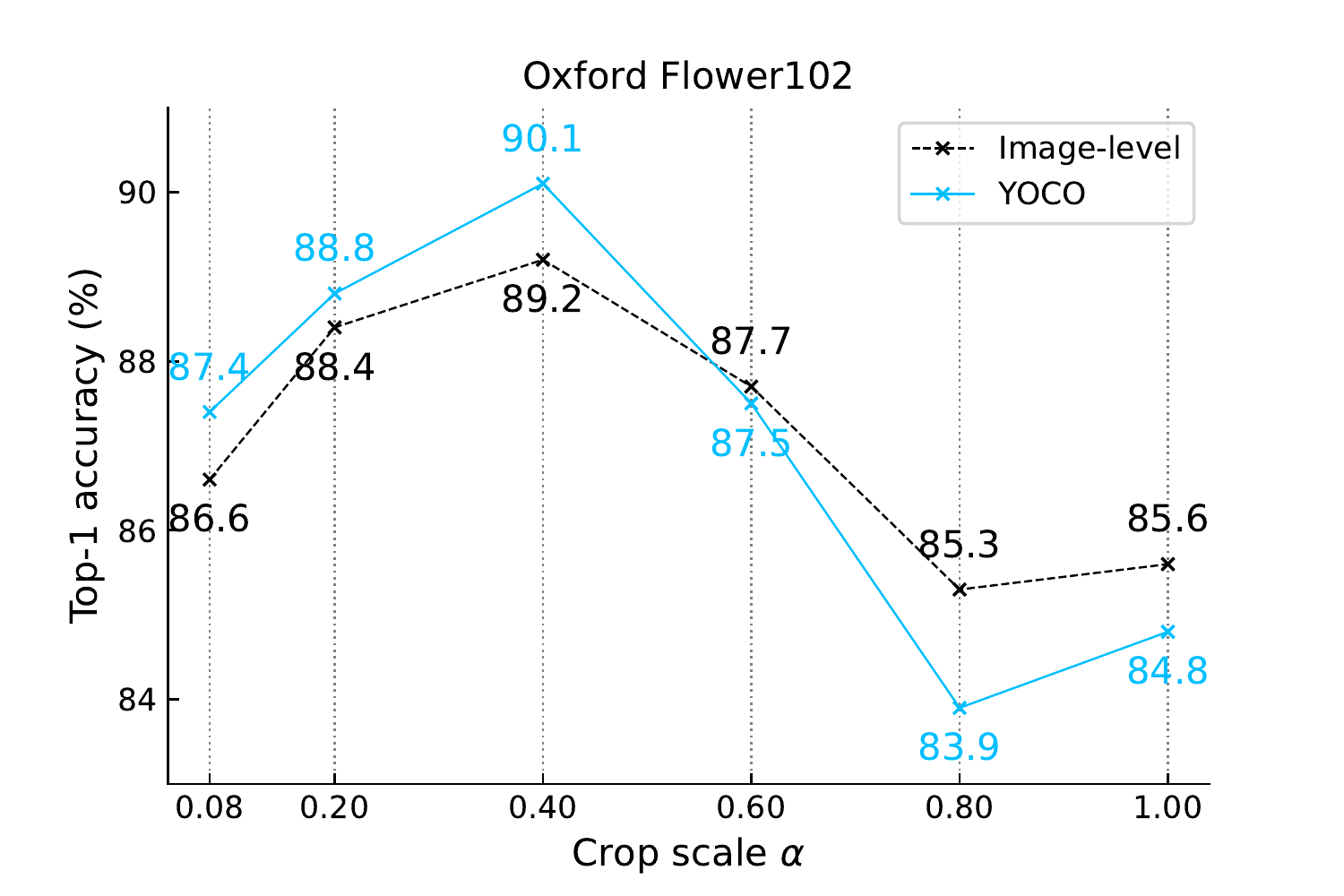}
  \end{minipage}   
    \begin{minipage}[t]{0.49\linewidth} 
    \centering 
\includegraphics[width=4.9cm]{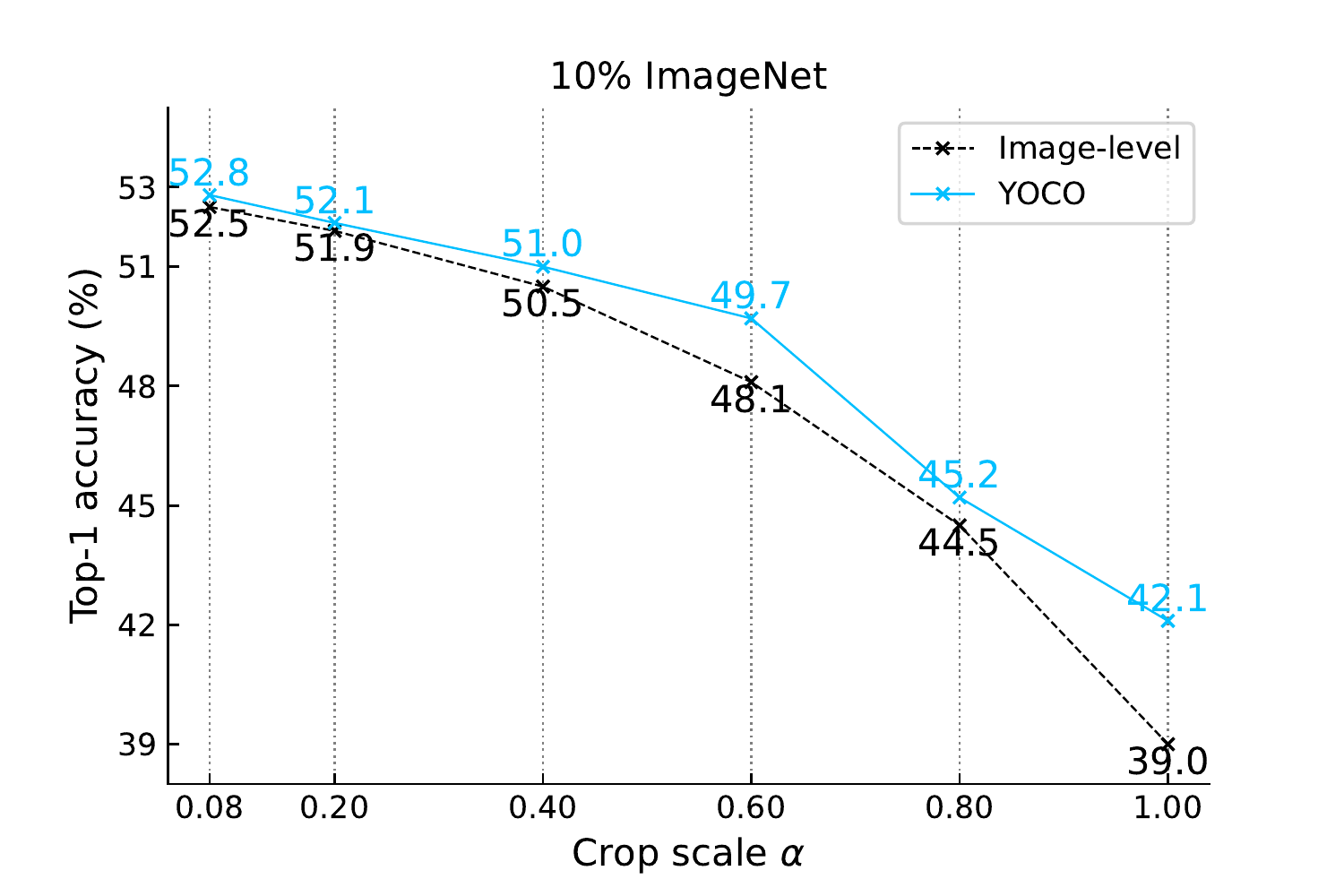}
  \end{minipage}   
    \vspace{-3mm}
     \caption{\textbf{When to employ YOCO?} Results of  \texttt{Horizontal flip} performed with image-level augmentation and YOCO on Oxford Flower102 (left) and 10\% ImageNet (right). We test multiple crop scale $(\alpha,1)$.  }
     \label{fig:when}
\end{figure}

To verify our hypothesis, we evaluate on Oxford Flower102~\cite{flower} (fine-grained flower classification dataset containing iconic flower images) and 10\% ImageNet. Results are presented in Figure~\ref{fig:when}. For iconic object images, YOCO is better than image-level augmentation only when a wider crop range is employed while YOCO always performs better on non-iconic object images. Thereby we conclude that for most augmentations, YOCO can outperform image-level augmentation, but for \texttt{Horizontal flip} YOCO should be employed depending on the properties and crop ratio of the dataset.

\subsection{Why Should We Employ YOCO?}
In the real-world environment, it is impossible to guarantee complete images as the source of inputs. That said, for a robust vision system applied in real-world practice, having the ability to recognize partial patterns is a must. In common practice, neural networks mostly gain such ability from the \texttt{random resized crop}. Applying YOCO to multiple augmentations further equips neural networks to have stronger partial recognition abilities, which also result in better holistic level recognition abilities for the neural networks.

YOCO is a booster for training more robust vision systems that can be applied in varying complex real scenarios. YOCO also serves as a free plug-and-play for various vision tasks, where no extra computation costs are brought, no parameters need to be manually selected, and no complicated implementation is required.

\subsection{Does YOCO Help Other Vision Tasks?}
YOCO boosts data augmentation in multiple high-level vision tasks, trained from scratch, or transferred from contrastive pre-training. We thus study whether YOCO helps other vision tasks, where we employ image deraining~\cite{Zamir2021MPRNet,han2021bid} and image super-resolution~\cite{yoo2020rethinking}, two representative low-level vision tasks, to further verify the generality of YOCO. 

\textbf{Image deraining:} We include two augmentations, \texttt{Horizontal flip} and \texttt{Vertical flip}, along with YOCO, to the training process of MPRNet~\cite{Zamir2021MPRNet}. Besides the augmentations, MPRNet is trained with the identical dataset (synthetic rain dataset~\cite{Zamir2021MPRNet} with 13712 pairs) and training settings. Results are presented in Table~\ref{tab:rain}. Our gain over the baseline is \textbf{0.73dB} on the Rain100L test set. 
\input{tables/rain}

\textbf{Image super-resolution:}
Following \texttt{CutBlur}~\cite{yoo2020rethinking}, we evaluate YOCO applied to mixture of augmentations (\texttt{Cutout}, \texttt{CutMix}, \texttt{Mixup}, \texttt{CutMixup}, \texttt{RGB perm}, \texttt{Blend}, and \texttt{CutBlur}) on two common datasets, DIV2K~\cite{agustsson2017ntire} and RealSR~\cite{cai2019toward}. All experiments are conducted using the CARN~\cite{ahn2018fast} model, where the training settings follows \texttt{CutBlur} but without pre-training on DIV2K in $\times$ 2 scale. In the synthetic training setting, models trained on the DIV2K dataset are evaluated on multiple test sets.
Table~\ref{tab:superres} presents the results, where
YOCO consistently outperforms image-level augmentation regardless of test sets in the synthetic setting, and shows on par performance in the realistic setting. 
\input{tables/superres}

We believe that our results on low-level vision further validate the generality and robustness of YOCO. 

\section{Limitation}
Though thorough empirical results show the superiority and generality of YOCO, the theoretical foundation can be further developed for shedding light on the interpretability of YOCO. Additionally, applying YOCO can be harmful to certain classes in the supervised classification task, as the effect of data augmentation is class dependent~\cite{balestriero2022effects}. To achieve superior classification performance, designing a class-specific strategy of applying YOCO is desirable.

\section{Conclusion}
We propose YOCO to boost augmentations, where YOCO has shown positive results in a variety of computer vision tasks and datasets. The competitiveness of our minimalist YOCO suggests that performing data augmentations can be a core component in training neural networks. We hope our study will attract the community’s attention in revisiting how to perform data augmentations. 

\section*{Acknowledgement}
We would like to thank the anonymous reviewers for their constructive feedback. 

\bibliography{main}
\bibliographystyle{icml2022}

\newpage
\appendix
\onecolumn
\section{Augmentations}
\label{sec:appendix}
For all studied augmentations, we list a brief description and implementation details here. For YOCO, we do not modify the setting of augmentation operations (probability, magnitude, \etc{}) unless specified. We employ the Pytorch/Torchvision implementation for most augmentations. 

\subsection{Geometric transformation}
\textbf{Horizontal flip.} Flip the image in the horizontal direction. We apply \texttt{Horizontal flip} with a probability of 0.5 as our default training setting.

\textbf{Vertical flip.} Flip the image in the vertical direction. We apply \texttt{Vertical flip} with a probability of 0.5.

\subsection{Photometric transformation}

\textbf{Color jitter.} 
Adjust the brightness, contrast, saturation, and hue of the image. We set $brightness = 0.4$, $contrast = 0.4$, $saturation = 0.4$, and $hue = 0.1$. The probability of being applied is 0.5. 

\textbf{Gaussian blur.} 
Blur the image with \texttt{Gaussian blur}. We set $sigma=(0.1, 2.0)$, kernel size as approximately 10\% of the image size, that is, for CIFAR images (32 $\times$ 32 resolution), a kernel size of 3 for image-level augmentation and kernel sizes of either 1 or 3 for YOCO. We apply \texttt{Gaussian blur} with a probability of 0.5.

\subsection{Information dropping}

\textbf{Random erasing.} 
Select a rectangular region in the image and erases its pixels. For CIFAR, we employ $scale=(0.02, 0.4)$, $ratio=(0.3, 3.3)$, and $value=0$. For ImageNet, we change $scale$ from $(0.02, 0.4)$ to $(0.02, 0.33)$. \texttt{Random erasing} is applied with a probability of 0.5.

\textbf{Cutout.} 
Drops the pixels of a square region in the image. The mask size for \texttt{Cutout} is set to 25\% of the image size (8 $\times$ 8 for CIFAR, 56 $\times$ 56 for ImageNet) and the location for dropping out is uniformly sampled. The dropped pixel is filled with 0 pixel. The probability of being applied is 0.5.

\subsection{Search-based}

\textbf{AutoAug.} 
Consists of 25 sub-policies (augmentations) searched from a pre-defined search space. Sub-policies span a wide range of augmentations. We employ the policy of CIFAR-10 and the policy of ImageNet for training on CIFAR datasets and ImageNet, respectively.

\textbf{RandAug.} 
Consists of searched augmentations from a reduced search space, where the magnitude strength of augmentations is controllable. RandAug can be applied uniformly across different tasks. For both CIFAR and ImageNet, we leverage an identical setting: 2 augmentation operations and a magnitude of 9. 

\subsection{Mix-based}

\textbf{Mixup.} 
Produce an element-wise convex combination of two images. The label of training images changes with the mixing ratio and mixing images. We employ \texttt{Mixup} with a probability of 1, and a mixing ratio randomly sampled from $Beta(1, 1)$ for both CIFAR and ImageNet. For YOCO, the mixing image is identical for each piece, but with different mixing ratios. This way produces a slightly better performance than involving two mixing images. When YOCO is applied, the ImageNet \textit{best} top-1 accuracy is \textbf{77.81} for one mixing image while being \textbf{77.72} for two mixing images. 

\textbf{CutMix.} 
Combine \texttt{Mixup} and \texttt{Cutout}, randomly crops a patch from the mixing image, and pastes it into the corresponding position of the training image. Similar to \texttt{Mixup}, we employ \texttt{CutMix} with a probability of 1, and draw the mixing ratio from $Beta(1, 1)$ for both the CIFAR and ImageNet. We also find one mixing image outperforms two mixing images. For PreResNet18 on CIFAR-100 dataset, the top-1 accuracy of one mixing image and two mixing images are \textbf{80.70} and \textbf{79.73}, respectively.

For group of augmentations, the setting of each augmentation is identical to that augmentation applied individually.

\section{Classification} 
\textbf{Standard deviation.} 
Table~\ref{tab:std} reports the standard deviation of applying YOCO, and not applying YOCO, on every single model and every single dataset in classification tasks. Overall, image-level augmentation is more stable than YOCO on CIFAR-100. However, on CIFAR-10 and ImageNet, the stability of YOCO is on par to Image-level augmentation.
\input{tables/std}

\section{Contrastive Learning} 
\label{sec:appendix_contrastive}
\subsection{Implementation details}
\textbf{Augmentations.} 
We follow the augmentations used in MoCo v2 and SimSiam. The augmentation procedure includes:  \texttt{Random resized crop} with scale in $[0.2,1.0]$,  \texttt{Color jitter} with $brightness = 0.4$, $contrast = 0.4$, $saturation = 0.4$, $hue = 0.1$, and being applied with a probability of 0.5, \texttt{Grayscale} with a probability of 0.8, \texttt{Guussian blur} with $sigma=(0.1, 2.0)$ and is applied with a probability of 0.5. For convenience, we move \texttt{Horizontal flip} to the end, which is applied with a probability of 0.5.

\textbf{Pre-training.} 
We follow the training setting of MoCo v2 and SimSiam. 

MoCo v2 is trained for 200 epochs with SGD as the optimizer. We use a weight decay of 0.0001 and a momentum of 0.9. The batch size is 256 and is distributed over 8 GPUs, and the initial learning rate is 0.03. The learning rate is multiplied by 0.1 at 120 and 160 epochs. The temperature is set to 0.2.

SimSiam is trained for 100 epochs only. The optimizer is SGD with a momentum of 0.9. The weight decay is 0.0001. We train with 8 GPUs and a batch size of 256. The initial learning rate is scaled to 0.05 with a cosine decay schedule. 

\textbf{Linear protocol.}
The linear protocol training setting follows MoCo v2 and SimSiam.

For MoCo v2, we train the linear classifier for 100 epochs. The initial learning rate is 30 and is multiplied by 0.1 at epochs 60 and 80. SGD optimizer is with a momentum of 0.9. The weight decay is 0. The linear classifier is trained with 8 GPUs and a batch size of 256.

SimSiam employs LARS optimizer~\cite{you2017large}. The linear classifier layer is trained for 90 epochs, and the initial learning rate is 0.2, which is scheduled with cosine decay. The weight decay is 0 and the momentum is 0.9. The batch size is 4096 and the training is distributed over 8 GPUs.

\textbf{Fine-tuning on ImageNet.} 
We follow the semi-supervised fine-tuning setting used in Jigsaw Clustering~\cite{chen2021jigsaw}. 

Pre-trained weights are fine-tuned for 100 epochs using SGD with a momentum of 0.9 as the optimizer. The batch size is 256 and we employ 4 GPUs for training. We do not apply any weight decay. The learning rate is multiplied by 0.1 at 30, 60, and 90 epochs.
For 10\% label fraction, we set the trunk learning rate to 0.01 while the last layer learning rate is set to 0.2. For 1\% label fraction, the learning rate for the trunk is 0.02 and is 5 for the final layer. 

\section{Visualization}
\label{sec:appendix_visualization}

\begin{figure*}[!htb]
     \centering
     \includegraphics[width = 17cm]
     {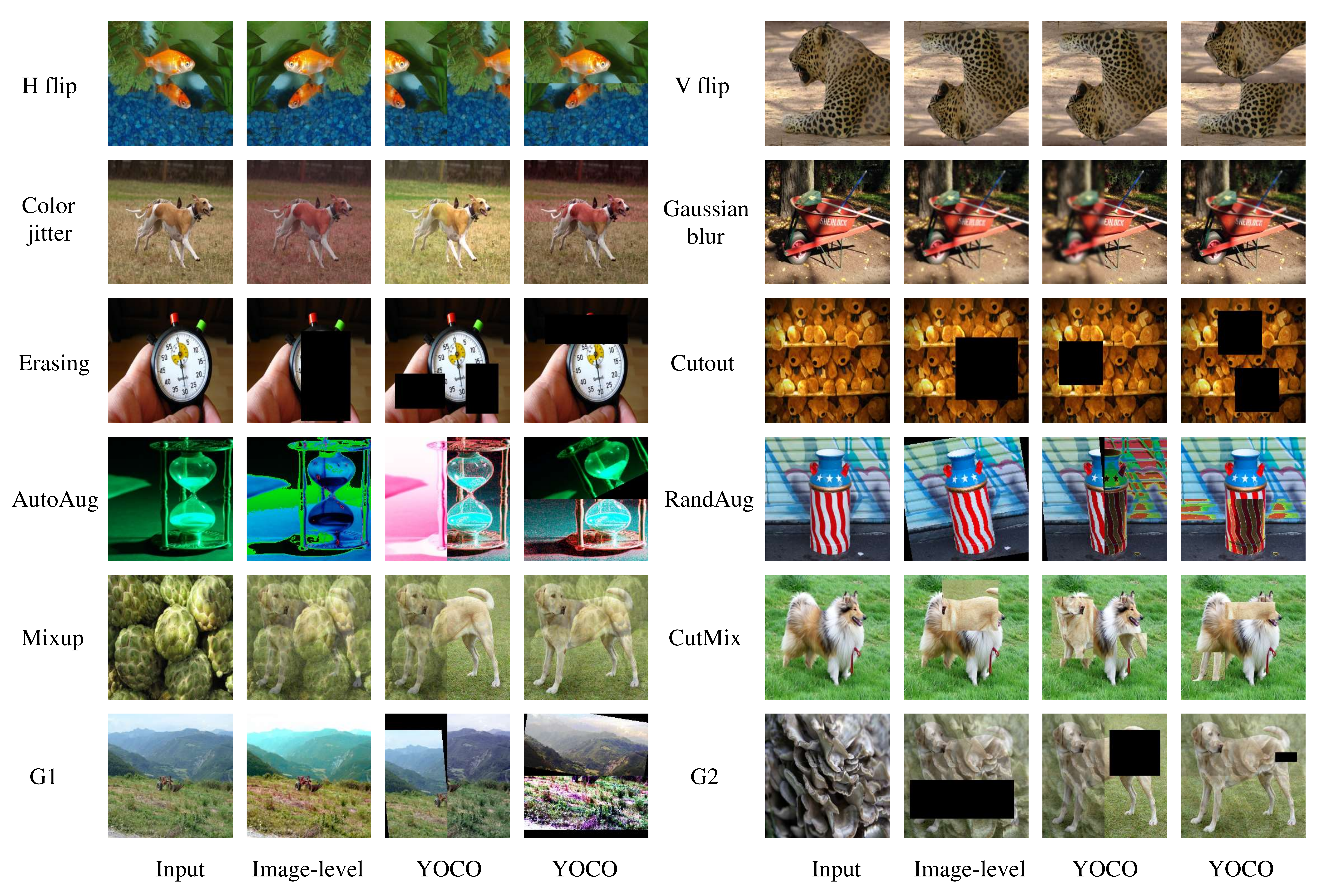}
     \caption{\textbf{Augmented images.} Randomly selected samples on ImageNet validation images. For each quadruplet, we show the original input image, augmented image from image-level augmentation, and two images from different cut dimensions produced by YOCO.  }
     \label{fig:visu}
\end{figure*}

\textbf{Augmented images.} We show uncurated random samples on ImageNet. Each quadruplet presents the original image, an augmented image produced by image-level augmentation, and two augmented images created by YOCO. Visualizations are presented in Figure~\ref{fig:visu}.

\begin{figure*}[!htb]
     \centering
     \includegraphics[width = 17cm]
     {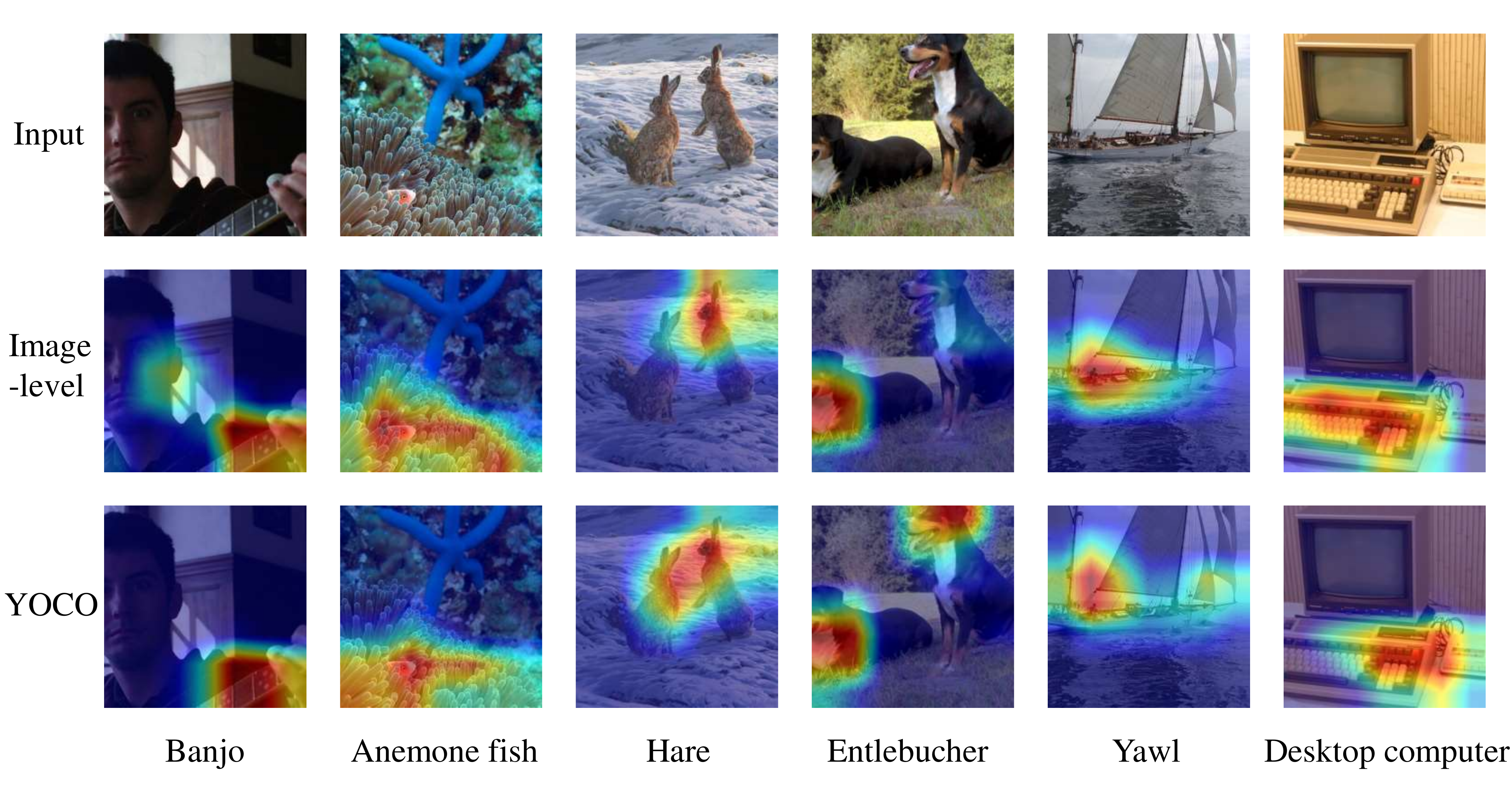}
     \caption{\textbf{Grad-CAM visualization}. We present the Grad-CAM results of models that are trained with image-level augmentation and YOCO. }
     \label{fig:grad}
\end{figure*}

\textbf{CAM visualization.} We compute Grad-CAM~\cite{selvaraju2017grad} for ResNet50 trained with YOCO and image-level augmentation. We take \texttt{Horizontal flip} as an example. Visualizations are presented in Figure~\ref{fig:grad}. For these samples, when only partial information is available (Banjo, Anemone fish), YOCO tends to focus more on that partial region instead of larger spatially distributed regions. YOCO also attends to focus on multiple structurally comprehensive regions, as shown in Hare,  Entlebucher, and Yawl, where YOCO noticed two objects while image-level pays attention to a certain one. However, YOCO sometimes fails to comprehensively localize class-related cues, as presented in Desktop computer. 

\begin{figure*}[!htb]
     \centering
     \includegraphics[width = 17cm]
     {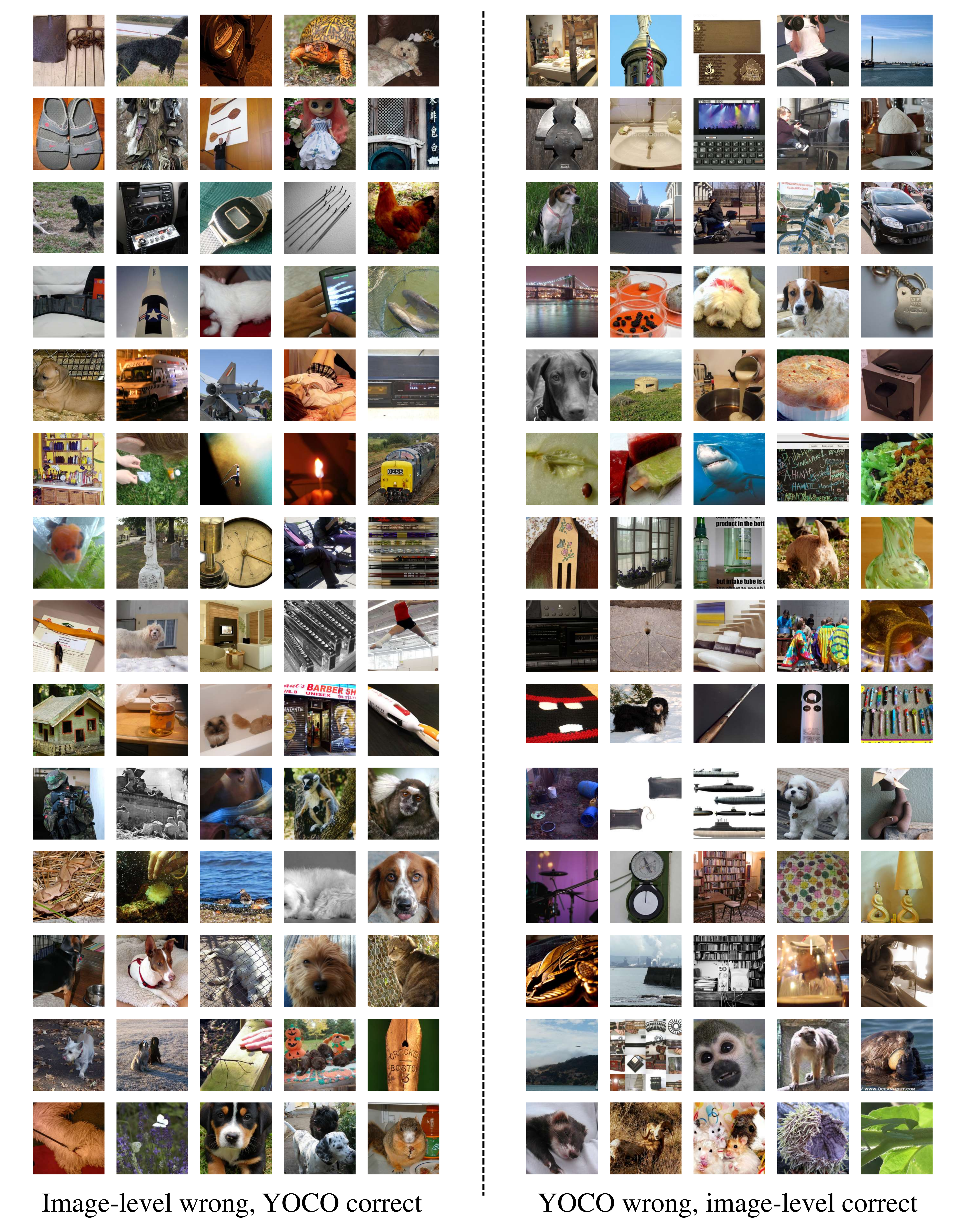}
     \caption{\textbf{Differently classified images.} Uncurated ImageNet validation images that are differently classified by models trained with YOCO and image-level augmentation.}
     \label{fig:gallery}
\end{figure*}

\textbf{Differently classified images.} To better understand the difference between image-level augmentation and YOCO. We present two collections containing randomly selected images that are differently classified by classifiers trained with YOCO and image-level augmentation. All images are taken from ImageNet validation set and images with private information are not presented. YOCO and image-level augmentation are both applied to \texttt{Horizontal flip}. Visualizations are presented in Figure~\ref{fig:gallery}. The set on the left side (image-level wrong, YOCO correct) contains more images that are with partial information only. 

\end{document}

%% file: tables/cifar10.tex
\begin{table*}[ht!]
  \centering
  \fontsize{8}{1}
  \selectfont
    \begin{tabular}{lc|c|c|c|c|c|c|c|c|c|c|c}
    \toprule
     \Rows{Models}&\multicolumn{2}{c|}{Geometric trans}&\multicolumn{2}{c|}{Photometric trans}&\multicolumn{2}{c|}{Information dropping}&\multicolumn{2}{c|}{Search-based}&\multicolumn{2}{c|}{Mix-based} &\multicolumn{2}{c}{Group} \cr
    &H flip &V flip & Jitter& Blur & Erasing & Cutout & AutoAug & RandAug &  Mixup & CutMix & G1 & G2\cr
    \midrule
    \gc{PreResNet18}& \gc{4.64} &  \gc{5.90} &  \gc{4.79} &  \gc{5.29} &  \gc{4.32} &  \gc{4.66} &  \gc{3.71} &  \gc{4.18} &  \gc{3.23} & \gc{3.45}  &\gc{3.69} & \gc{3.63} \cr
    + YOCO& 5.05  & 5.52  & 4.70 & 5.00 & 4.28  & 4.36 & 3.38 & 4.08 & 3.11 & 3.38  & 3.56 & 3.25\cr
    $\triangle$ & \re{+0.41}  & \bl{-0.38} & \bl{-0.09} & \bl{-0.29} & \bl{-0.04}  & \bl{-0.30} & \bl{-0.33} & \bl{-0.10} & \bl{-0.12} & \bl{-0.07} &\bl{-0.13}& \bl{-0.38} \cr
    \midrule
    \gc{Xception}& \gc{4.62} &  \gc{5.20} &  \gc{4.52} &  \gc{5.31} &  \gc{4.07} &  \gc{4.15} &  \gc{3.56} &  \gc{4.01} &  \gc{3.70} & \gc{3.66} & \gc{4.04} &\gc{3.41}  \cr
    + YOCO& 4.91 & 5.11 & 4.51  & 4.65 & 4.07 &  3.94 &  3.31 & 3.95 & 3.17 & 3.22   &3.32 &3.20\cr
    $\triangle$ & \re{+0.29}  & \bl{-0.09}  & \bl{-0.01}  & \bl{-0.66} & 0.00 & \bl{-0.21} & \bl{-0.25} & \bl{-0.06} & \bl{-0.53} & \bl{-0.44} & \bl{-0.72} & \bl{-0.21}  \cr
    \midrule    
    \gc{DenseNet121}& \gc{4.58} &  \gc{5.21} &  \gc{4.72} &  \gc{5.09} &  \gc{4.17} &  \gc{4.55} &  \gc{3.83} &  \gc{4.22} &  \gc{3.71} & \gc{3.53}   & \gc{3.98} &\gc{3.57} \cr
    + YOCO&  5.09& 5.18 & 4.70  & 4.76& 4.13  & 4.25 & 3.67 & 3.91 & 3.28 & 3.50   & 3.50 &3.36 \cr
    $\triangle$ & \re{+0.51}  & \bl{-0.03}  & \bl{-0.02}  & \bl{-0.33} & \bl{-0.04} & \bl{-0.30} & \bl{-0.16} & \bl{-0.31} & \bl{-0.43} & \bl{-0.03}& \bl{-0.48}  & \bl{-0.21} \cr
    \midrule
    \gc{ResNeXt50}& \gc{4.69} &  \gc{5.12} &  \gc{4.74} &  \gc{5.66} &  \gc{4.18} &  \gc{4.93} &  \gc{3.65} &  \gc{4.09} &  \gc{3.69} & \gc{3.52}   & \gc{3.79} &\gc{3.66}  \cr
    + YOCO& 5.10 & 5.03  & 4.60  &  5.27 & 3.96  & 4.64 & 3.26 & 3.91 & 3.45 & 3.44  &3.26 & 3.41\cr
    $\triangle$ & \re{+0.41}  & \bl{-0.09}  & \bl{-0.14}  & \bl{-0.39} &  \bl{-0.22} & \bl{-0.29} & \bl{-0.39} & \bl{-0.18} & \bl{-0.24} & \bl{-0.08}  & \bl{-0.53} & \bl{-0.25}\cr
    \midrule
    \gc{WRN-28-10}& \gc{3.51} &  \gc{4.10} &  \gc{3.63} &  \gc{3.99} &  \gc{2.92} &  \gc{3.26} &  \gc{2.72} &  \gc{3.23} &  \gc{2.60} & \gc{2.86}  & \gc{2.75} &\gc{2.79} \cr
    + YOCO& 3.53 & 4.09 & 3.44  &3.97 & 2.89 & 2.96 & 2.62 &3.05 & 2.46 & 2.82   & 2.65 &2.35 \cr
    $\triangle$ & \re{+0.02}  & \bl{-0.01}  & \bl{-0.19}  & \bl{-0.02} &  \bl{-0.03} & \bl{-0.30} & \bl{-0.10} & \bl{-0.18} & \bl{-0.14} & \bl{-0.04} & \bl{-0.10} & \bl{-0.44} \cr
     \midrule 
         \gc{ViT}& \gc{4.69}&\gc{5.51} &\gc{4.74} & \gc{5.49} &\gc{4.32}&\gc{4.44} & \gc{3.76} & \gc{4.13} & \gc{3.67} & \gc{3.49}  & \gc{3.82}  & \gc{3.27}\cr
    + YOCO& 4.92 & 5.34 & 4.70 & 4.95 & 4.18& 4.42& 3.54 & 3.96 & 3.17  & 3.21 & 3.51 & 3.14 \cr
    $\triangle$ & \re{+0.23} & \bl{-0.17}&\bl{-0.04} & \bl{-0.54}  & \bl{-0.14} & \bl{-0.02 }& \bl{-0.22} & \bl{-0.17} & \bl{-0.50}  & \bl{-0.28} & \bl{-0.31}  & \bl{-0.13} \cr
    \midrule    
    \gc{Swin}& \gc{4.83}&\gc{5.27} &\gc{4.90} & \gc{5.15} &\gc{4.26}&\gc{4.55} & \gc{3.63} & \gc{4.15} & \gc{3.56} & \gc{3.64}  & \gc{3.74}  & \gc{3.58}\cr
    + YOCO& 4.85 & 5.25 & 4.40 & 5.12  & 4.23 & 4.03& 3.54 &  4.13 & 3.16  & 3.22 & 3.44 & 3.46 \cr
    $\triangle$ & \re{+0.02} & \bl{-0.02} & \bl{-0.50} & \bl{-0.03}  & \bl{-0.03} & \bl{-0.52} & \bl{-0.09} & \bl{-0.02}  & \bl{-0.40}  & \bl{-0.42} & \bl{-0.30}  & \bl{-0.12} \cr
    \midrule  
    Average $\triangle$ & \re{+0.27}  & \bl{-0.11}  & \bl{-0.14} & \bl{-0.32} & \bl{-0.07} & \bl{-0.28} & \bl{-0.22} & \bl{-0.15} & \bl{-0.34}& \bl{-0.19} & \bl{-0.37} & \bl{-0.25}  \cr
    \bottomrule
    \end{tabular}
     \caption{\textbf{Test top-1 error rate on CIFAR-10}. We evaluate 12 augmentations on 7 network architectures. Among 84 results, 76 of them are improved by applying YOCO. Improved results are highlighted in \bl{blue}. Best viewed in color. Standard deviations lie in $(0.04, 0.15)$. Average $\triangle$ is the difference between YOCO and image-level augmentation, which is based on the mean of 21 independent runs in total.  }
     \label{tab:cifar10}
\end{table*}

%% file: tables/cifar100.tex
\begin{table*}[ht!]
  \centering
  \fontsize{8}{1}
  \selectfont
    \begin{tabular}{lc|c|c|c|c|c|c|c|c|c|c|c}
    \toprule
     \Rows{Models}&\multicolumn{2}{c|}{Geometric trans}&\multicolumn{2}{c|}{Photometric trans}&\multicolumn{2}{c|}{Information dropping}&\multicolumn{2}{c|}{Search-based}&\multicolumn{2}{c|}{Mix-based} &\multicolumn{2}{c}{Group} \cr
    &H flip &V flip & Jitter& Blur & Erasing & Cutout & AutoAug & RandAug &  Mixup & CutMix & G1 & G2\cr
    \midrule
    \gc{PreResNet18}& \gc{23.36}&\gc{25.16} &\gc{23.95} & \gc{25.96} &\gc{24.03}&\gc{25.01} & \gc{21.94} & \gc{22.70} & \gc{21.32} & \gc{20.05}  & \gc{22.97}  & \gc{21.23}\cr
    + YOCO& 24.66&24.73 &23.62 & 25.11 &23.94 &23.41& 20.55 & 22.70 & 19.62 &19.30 & 21.06 & 20.92 \cr
    $\triangle$ & \re{+1.30} & \bl{-0.43}  & \bl{-0.33} & \bl{-0.85} & \bl{-0.09}  & \bl{-1.60} & \bl{-1.39} & 0.00 & \bl{-1.70} & \bl{-0.75}  & \bl{-1.91} & \bl{-0.31} \cr
    \midrule
    \gc{Xception}& \gc{23.34}&\gc{25.16} &\gc{23.39} & \gc{26.17} &\gc{25.01} &\gc{25.44} & \gc{22.50} & \gc{22.87} & \gc{21.41} & \gc{20.17} & \gc{23.02} & \gc{21.79}  \cr
    + YOCO& 24.46&24.65 &23.42 & 25.51 &23.74 &23.56 & 20.29 & 22.45 & 19.96  &19.25 &20.75  &20.68 \cr
    $\triangle$ & \re{+1.12} & \bl{-0.51}  & \re{+0.03} & \bl{-0.66} & \bl{-1.27} & \bl{-1.88} & \bl{-2.21} & \bl{-0.42} & \bl{-1.45} & \bl{-0.92} & \bl{-2.27} & \bl{-1.11}  \cr
    \midrule    
    \gc{DenseNet121}& \gc{23.14}&\gc{23.71} &\gc{24.25} & \gc{26.21} &\gc{23.97} &\gc{24.71} & \gc{22.43} & \gc{22.66} & \gc{21.17} &\gc{19.50}& \gc{22.80}  & \gc{21.23}  \cr
    + YOCO& 24.21&24.62 &23.14 & 24.68 &23.34 &23.83 & 20.63 & 23.11 & 19.87 & 19.26  & 21.13 & 20.95\cr
    $\triangle$ & \re{+1.07}  & \re{+0.91}  & \bl{-1.11}  & \bl{-1.53} & \bl{-0.63}  & \bl{-0.88} & \bl{-1.80} & \re{+0.45} & \bl{-1.30} & \bl{-0.24} & \bl{-1.67} & \bl{-0.28}   \cr
    \midrule
    \gc{ResNeXt50}& \gc{24.60}&\gc{24.61} &\gc{23.58} & \gc{25.86} &\gc{24.50} &\gc{24.84} & \gc{22.32} & \gc{22.77} & \gc{20.78} &\gc{19.72} & \gc{22.69} & \gc{21.69}  \cr
    + YOCO& 25.11&24.22 &23.76 & 25.66 &23.51 &23.77 & 21.01 & 22.47 & 19.74 & 19.68 & 20.58 & 20.55\cr
    $\triangle$ & \re{+0.51}  & \bl{-0.39}  & \re{+0.18}  & \bl{-0.20} & \bl{-0.99}  & \bl{-1.07} & \bl{-1.31} & \bl{-0.30}& \bl{-1.04}& \bl{-0.04}  & \bl{-2.11} & \bl{-1.14}\cr
    \midrule
     \gc{WRN-28-10}& \gc{19.90}&\gc{21.70} &\gc{19.54} & \gc{21.95} &\gc{19.79} &\gc{19.73} & \gc{18.07} & \gc{19.25} & \gc{18.93} &\gc{17.03} & \gc{18.39} & \gc{17.74} \cr
    + YOCO& 20.44&21.34 &19.59 & 21.17 &19.65 &19.68 & 17.17 & 19.29 & 17.64 & 17.29 & 18.89 & 17.72 \cr
    $\triangle$ & \re{+0.54}  & \bl{-0.36}  & \re{+0.05} & \bl{-0.78} & \bl{-0.14} & \bl{-0.05} & \bl{-0.90} & \re{+0.04} & \bl{-1.29}& \re{+0.26} & \bl{-0.50} & \bl{-0.02} \cr
    \midrule
    \gc{ViT}& \gc{23.67}&\gc{24.01} &\gc{24.53} & \gc{26.06} &\gc{24.09}&\gc{25.10} & \gc{22.14} & \gc{22.79} & \gc{21.15} & \gc{20.27}  & \gc{22.75}  & \gc{21.64}\cr
    + YOCO& 24.52& 24.74 &23.49 & 25.04  & 23.20 & 23.38 & 20.61 & 22.34 & 19.26  & 19.17 & 20.85 & 21.20 \cr
    $\triangle$ & \re{+0.85} & \re{+0.73} & \bl{-1.04} & \bl{-1.02}  & \bl{-0.89} & \bl{-1.72} & \bl{-1.53} & \bl{-0.45}  & \bl{-1.89}  & \bl{-1.10} & \bl{-1.90}  & \bl{-0.44} \cr
    \midrule    
    \gc{Swin}& \gc{23.57}&\gc{24.51} &\gc{24.04} & \gc{25.49} &\gc{24.81}&\gc{25.11} & \gc{22.30} & \gc{23.14} & \gc{21.42} & \gc{19.78}  & \gc{22.92}  & \gc{21.26}\cr
    + YOCO& 24.66 & 24.37 & 23.96 & 25.39  & 23.22 & 23.86 & 20.70 & 22.62 & 19.43 & 19.74 & 20.75 & 20.88 \cr
    $\triangle$ & \re{+1.09} & \bl{-0.14} & \bl{-0.08} & \bl{-0.10} & \bl{-1.59} & \bl{-1.25}& \bl{-1.60} & \bl{-0.52} & \bl{-1.99}  & \bl{-0.04} & \bl{-2.17}  & \bl{-0.38} \cr
    \midrule
    Average $\triangle$ & \re{+0.93}  & \bl{-0.03}  & \bl{-0.33} & \bl{-0.73} & \bl{-0.80} & \bl{-1.21} & \bl{-1.53} & \bl{-0.17} & \bl{-1.52}& \bl{-0.40} & \bl{-1.79} & \bl{-0.53} \cr     
    \bottomrule
    \end{tabular}
     \caption{\textbf{Test top-1 error rate on CIFAR-100}, where 12 augmentations and 7 network architectures are evaluated. YOCO outperforms image-level augmentation in 68 results out of 84 results. \bl{blue} indicates improved results. Best viewed in color. Standard deviations are between $(0.09, 0.41)$. Average $\triangle$ is the difference between YOCO and image-level augmentation over 21 independent trials. }
     \label{tab:cifar100}
\end{table*}

%% file: tables/imagenet_robust.tex
\begin{table*}[!htbp]
  \centering
  \fontsize{7.5}{3}
  \selectfont
    \begin{tabular}{lc|c|c|c|c|c|c|c}
    \toprule
     \Rows{Augs}&\multicolumn{1}{c|}{Generalization}&\multicolumn{1}{c|}{Partial}&\multicolumn{1}{c|}{Calibration}&\multicolumn{2}{c|}{Adversarial attacks}&\multicolumn{2}{c|}{Corruptions} &\multicolumn{1}{c}{Distribution shift} \cr
     &Clean  &Clean  & Clean, RMS$\downarrow$ &FGSM attack &PGD attack  & Random replace & Gaussian noise & ImageNet-A \cr
    \midrule
    \gc{H flip} &  \gc{77.10/76.59} &  \gc{55.21/56.14} & \gc{6.42/8.81} & \gc{17.28/21.01}&\gc{9.37/14.98}  & \gc{58.90/58.77} &\gc{73.58/72.92} & \gc{3.39/4.29} \cr
    + YOCO &  77.28/77.01  &  56.27/56.44  &7.87/8.19 & 20.59/21.88 &14.21/15.21 & 59.11/58.91 &73.63/72.96 & 3.80/4.32 \cr
    $\triangle$ &  \bl{+0.18}/\bl{+0.42} &  \bl{+1.06}/\bl{+0.30} & \re{+1.45}/\bl{-0.62}  & \bl{+3.31}/\bl{+0.87} & \bl{+4.84}/\bl{+0.23} & \bl{+0.21}/\bl{+0.14} &  \bl{+0.05}/\bl{+0.04} & \bl{+0.41}/\bl{+0.03} \cr
    
    \midrule
    \gc{Jitter} &  \gc{77.15/76.87}  &  \gc{56.04/56.34} & \gc{8.06/8.19}  &  \gc{18.31/18.80} &  \gc{11.07/11.89} & \gc{59.11/58.82} & \gc{73.74/73.38}  & \gc{3.98/4.13} \cr
    + YOCO &  77.35/77.12 &  56.41/56.60 & 7.50/8.07& 16.32/17.77  &  5.96/6.90 & 59.40/59.32&  74.46/74.14 & 4.29/4.56\cr
    $\triangle$& \bl{+0.20}/\bl{+0.25} &  \bl{+0.37}/\bl{+0.26}  & \bl{-0.56}/\bl{-0.12}&  \re{-1.99}/\re{-1.03} & \re{-5.11}/\re{-4.99} &\bl{+0.29}/\bl{+0.50} &\bl{+0.72}/\bl{+0.76} &\bl{+0.31}/\bl{+0.43} \cr
    \midrule   
    \gc{Erasing} & \gc{77.40/77.09} & \gc{56.75/56.82} & \gc{7.67/7.88} & \gc{22.01/22.67} & \gc{12.74/13.31} & \gc{58.75/58.44} & \gc{73.81/73.48}  & \gc{4.11/4.32} \cr
    + YOCO  &  77.29/77.20 &  56.66/56.76 & 7.57/8.00 & 22.34/22.82 & 12.93/14.00 & 58.80/58.85 & 73.95/73.59 & 4.12/4.28 \cr
    $\triangle$ & \re{-0.11}/\bl{+0.11} &  \re{-0.09}/\re{-0.06} & \bl{-0.10}/\re{+0.12}&  \bl{+0.33}/\bl{+0.15}  & \bl{+0.19}/\bl{+0.69} & \bl{+0.05}/\bl{+0.41} & \bl{+0.14}/\bl{+0.11} & \bl{+0.01}/\re{-0.04}  \cr
    \midrule
    \gc{AutoAug} &  \gc{77.55/76.93} &  \gc{55.61/55.05} & \gc{6.17/6.73} & \gc{11.51/11.62} & \gc{2.05/2.18}  & \gc{58.36/58.12} & \gc{74.45/73.85}  & \gc{4.54/4.69} \cr
    + YOCO &  77.88/77.65 &  56.48/56.52   &  6.04/6.25 & 12.09/11.73 & 1.11/0.86 & 58.90/58.29 & 76.03/75.70 & 5.01/4.80 \cr
    $\triangle$ &\bl{+0.33}/\bl{+0.72} &\bl{+0.87}/\bl{+1.47} & \bl{-0.13}/\bl{-0.48}  &  \bl{+0.58}/\bl{+0.11}  & \re{-0.94}/\re{-1.32} & \bl{+0.54}/\bl{+0.17}& \bl{+1.58}/\bl{+1.85} & \bl{+0.47}/\bl{+0.11}  \cr
    \midrule
    \gc{Mixup} &  \gc{77.72/77.67} &  \gc{55.27/55.40}  & \gc{8.23/9.03}  & \gc{35.92/35.41} & \gc{7.90/7.44}  &  \gc{61.46/61.07} & \gc{75.47/75.12} & \gc{8.37/8.59} \cr
    + YOCO &  77.81/77.74  &  56.00/55.97  &4.04/4.42 &39.63/39.67  &13.02/13.63   &61.14/60.85 &75.10/74.88 & 8.27/8.55  \cr
    $\triangle$ & \bl{+0.09}/\bl{+0.07} &\bl{+0.73}/\bl{+0.57} & \bl{-4.19}/\bl{-4.61}&  \bl{+3.71}/\bl{+4.26} & \bl{+5.12}/\bl{+6.19} & \re{-0.32}/\re{-0.22} & \re{-0.37}/\re{-0.24} & \re{-0.10}/\re{-0.04}
    \cr
    \midrule
    \gc{G1} &  \gc{77.61/77.19}   &  \gc{55.57/55.24} & \gc{6.29/6.19} & \gc{13.11/13.22} & \gc{2.34/2.87} & \gc{58.94/58.99}  & \gc{75.22/74.94} & \gc{5.01/5.22} \cr
    + YOCO &  77.76/77.64 &  56.00/55.97 &  6.17/6.02 & 13.62/13.61 &  1.82/2.61 & 59.15/59.03  & 75.73/75.59 & 5.41/5.47 \cr
    $\triangle$ &\bl{+0.15}/\bl{+0.45} &\bl{+0.43}/\bl{+0.73} & \bl{-0.12}/\bl{-0.17}  &  \bl{+0.51}/\bl{+0.39}  & \re{-0.52}/\re{-0.26}  & \bl{+0.21}/\bl{+0.04} & \bl{+0.51}/\bl{+0.65}  & \bl{+0.40}/\bl{+0.25} \cr    
    \bottomrule
    \end{tabular}
     \caption{\textbf{Results of applying YOCO to train a ResNet50 on ImageNet}. We evaluate generalization (top-1 accuracy), partial image recognition (top-1 accuracy), calibration (RMS), robustness against adversarial attacks (top-1 accuracy), corruption robustness (top-1 accuracy), and robustness under distribution shifts (top-1 accuracy). We report both\textit{ best} result (left) and \textit{last} result (right).  Improved results are highlighted in \bl{blue} color. Best viewed in color. $\triangle$ is the difference between YOCO and image-level augmentation. }
     \label{tab:imagenet_robust}
\end{table*}

%% file: tables/contrastive_transfer.tex
\begin{table*}[!htbp]
  \centering
  \fontsize{8}{3}
  \selectfont
    \begin{tabular}{lc|c|c|c|c|c|c|c|c|c|c|cc}
    \toprule
    \Rows{Method}&\multicolumn{3}{c|}{ImageNet classification} &\multicolumn{3}{c|}{VOC detection} &\multicolumn{3}{c|}{COCO detection}  &\multicolumn{3}{c}{COCO instance seg} \cr
    & linear protocol&  1\% label & 10\% label  & AP$_\text{50}$ & AP & AP$_\text{75}$ & AP$_\text{50}$ & AP & AP$_\text{75}$ & AP$^\text{mask}_\text{50}$ & AP$^\text{mask}$ & AP$^\text{mask}_\text{75}$   \cr
    \midrule
    \gc{MoCo v2} & \gc{67.5}  & \gc{34.5} & \gc{61.1} & \gc{81.9} & \gc{56.8} & \gc{62.9}  & \gc{57.3}  & \gc{38.0} & \gc{40.8} & \gc{54.1} & \gc{33.3} & \gc{35.5}   \cr
    + YOCO & 67.6  & 34.9  & 61.5 & 82.4 & 56.6 & 63.6 & 57.4 & 38.1 & 41.6 & 54.1 & 33.4 & 35.5 \cr
    \midrule
    \gc{SimSiam} & \gc{68.1}  & \gc{17.1}  & \gc{57.3} & \gc{80.2} & \gc{54.5} & \gc{60.0} & \gc{51.4} & \gc{33.5} & \gc{36.1} & \gc{48.7} & \gc{29.9} & \gc{32.0}  \cr
    +YOCO & 68.3  & 17.3  & 57.7 & 80.2 & 54.5 & 60.1 & 53.1 & 34.6 & 37.3 & 50.1 & 30.8 & 32.9 \cr
    \bottomrule
    \end{tabular}
     \caption{\textbf{Results of contrasitve learning}. MoCo v2 and SimSiam are pre-trained for 200 and 100 epochs in ImageNet, respectively. All are based on ResNet50 pre-trained with two 224×224 views. The evaluation metric of ImageNet classification is Top-1 accuracy.  }
     \label{tab:contrastive}
\end{table*}

%% file: tables/contrastive_supp.tex
\begin{table}[!htbp]
  \centering
  \fontsize{6.3}{3}
  \selectfont
    \begin{tabular}{lc|c|c|c|c|cc}
    \toprule
    \Rows{Method}&\multicolumn{3}{c|}{ImageNet classification} &\multicolumn{3}{c}{VOC detection}  \cr
    & linear protocol&  1\% label & 10\% label  & AP$_\text{50}$ & AP & AP$_\text{75}$     \cr
    \midrule
\gc{MoCo v2} & \gc{67.5}  & \gc{34.5} & \gc{61.1} & \gc{81.9} & \gc{56.8} & \gc{62.9} \cr
    \midrule
    (1) & 67.6  & 34.9  & 61.5 & 82.4 & 56.6 & 63.6 \cr
    (2) & 67.3  & 34.8  & 61.3 & 82.5 & 56.9 & 63.4 \cr
    (3) & 67.2  & 34.5  & 61.2 & 82.4 & 56.6 & 62.7 \cr
    (4) & 67.0  & 34.4  & 61.0 & 82.1 & 56.0 & 61.9 \cr

    \bottomrule
    \end{tabular}
     \caption{\textbf{The effect of YOCO applied to multiple augmentations in contrastive learning}. We study YOCO applied to (1):  \texttt{Horizontal flip}, (2):  \texttt{Horizontal flip} +  \texttt{Gaussian blur}, (3):  \texttt{Horizontal flip} +  \texttt{Gaussian blur} +  \texttt{Grayscale}, and (4):  \texttt{Horizontal flip} +  \texttt{Gaussian blur} +  \texttt{Grayscale} +  \texttt{Color jitter}.  }
     \label{tab:contrastive_supp}
\end{table}

%% file: tables/position.tex
\begin{table}[!htbp]
  \centering
  \fontsize{8}{3}
  \selectfont
    \begin{tabular}{lc|c|c|c|c}
    \toprule
    \Rows{$\alpha$}&\multicolumn{5}{c}{Augmentations} \cr
    &H flip& Jitter& Erasing &AutoAug & Mixup \cr
    \midrule
    0.2 & 76.2  & 76.3  & 75.9 & 79.8 & 79.7\cr
    0.4 & 75.9  & 76.5  & 75.8 & 79.8 & 79.7\cr
    0.6 & 76.5  & 76.4  & 76.3 & 79.9  & 79.8\cr
    0.8 & 76.4  & 76.3  & 75.9 & 79.8 & 79.5\cr
    1.0 & 76.4  & 76.2  & 75.7 & 80.2 & 79.4\cr
    \midrule
    YOCO & 75.6  & 76.6  & 76.4 & 79.4 & 80.3\cr            
    \bottomrule
    \end{tabular}
     \caption{\textbf{Position of the cut}. Test top-1 accuracy on CIFAR-100 are reported. We randomly sample the position of the cut from a beta distribution $\mathrm{Beta}(\alpha, \alpha)$.}
     \label{tab:position}
\end{table}

%% file: tables/cutnumber.tex
\begin{table}[!htbp]
  \centering
  \fontsize{8}{3}
  \selectfont
    \begin{tabular}{lc|c|c|c|c}
    \toprule
    Number&\multicolumn{5}{c}{Augmentations} \cr
    H,W &H flip& Jitter& Erasing &AutoAug & Mixup \cr
    \midrule
    1,1 & 70.8  & 76.1 & 76.0 & 77.8 & 79.4\cr
    1,2 & 69.1  & 75.9  & 76.0 & 77.2 & 79.2\cr
    1,3 & 68.0  & 75.5  & 75.9 & 76.7  & 79.2\cr
    2,1 & 69.7  & 76.5  & 75.9 & 76.9 & 80.1\cr
    2,2 & 68.7  & 75.9  & 75.6 & 76.5 & 79.6\cr
    2,3 & 67.9  & 75.7  & 75.6 & 76.2 & 79.5\cr
    3,1 & 69.9  & 75.9 &  75.8 & 77.0 & 80.2\cr
    3,2 & 67.6  & 75.8  & 75.5 & 76.3 & 79.8\cr
    3,3 & 67.4  & 75.2  & 75.5 & 76.0 & 79.6\cr    
    \midrule
    YOCO & 75.6  & 76.6  & 76.4 & 79.4 & 80.3\cr            
    \bottomrule
    \end{tabular}
     \caption{\textbf{Number of cuts}. We report test top-1 accuracy on CIFAR-100. The number of cuts are set to 1, 2, or 3 for both height dimension and width dimension.}
     \label{tab:cutnumber}
\end{table}

%% file: tables/solely.tex
\begin{table}[!htbp]
  \centering
  \fontsize{8}{3}
  \selectfont
    \begin{tabular}{lc|c|c|c}
    \toprule
    \Rows{Combination}&\multicolumn{4}{c}{Augmentations} \cr
    &H flip& Jitter& Erasing & Mixup \cr
    \midrule
    Original & 72.7 & 76.8 & 76.8  & 76.8\cr
    Fully-aug & 71.7 & 76.2 & 76.5 & 78.8 \cr
    Partially-aug & 74.2 & 76.5 & 76.5 & 79.6\cr
    Diversified fully-aug & 72.5 & 76.3 & 76.9 & 79.7 \cr
    \bottomrule
    \end{tabular}
     \caption{\textbf{How does YOCO work?} Test top-1 accuracy on CIFAR-100. We train image classifiers with one kind of data solely. }
     \label{tab:solely}
\end{table}

%% file: tables/rain.tex
\begin{table}[!htbp]
  \centering
  \fontsize{6}{3}
  \selectfont
    \begin{tabular}{lcccccc}
    \toprule
    Methods & Test100 &  Rain100H & Rain100L  & Test2800 & Test1200 & Average \cr
    \midrule
    \gc{MPRNet} & \gc{30.27} & \gc{30.41} & \gc{36.40} & \gc{33.64} & \gc{32.91} & \gc{32.73} \cr
    + YOCO  & 30.33 & 30.53 & 37.13 & 33.64 & 32.84 & 32.89\cr
    \bottomrule
    \end{tabular}
    \vspace{-3mm}
    \caption{\textbf{Image deraining}. PSNR results of YOCO applied to image deraining tasks. We choose the state-of-the-art MPRNet as our baseline. Results are further improved with YOCO.}
    \label{tab:rain}
\end{table}

%% file: tables/superres.tex
\begin{table}[!htbp]
  \centering
  \fontsize{7}{3}
  \selectfont
    \begin{tabular}{lc|c|c|c|cc}
    \toprule
    \Rows{Aug}&\multicolumn{4}{c|}{Synthetic} &\multicolumn{1}{c}{Realistic } \cr
    & DIV2K&  Set14 & Urban100  & Manga109 & RealSR \cr
    \midrule
    \gc{No aug} & \gc{28.83}  & \gc{28.49} & \gc{25.82} & \gc{30.11} & \gc{28.78} \cr
    \gc{+ Image-level} & \gc{28.83}  & \gc{28.48} & \gc{25.82} & \gc{30.08} & \gc{28.99} \cr
    + YOCO & 28.84 & 28.50 & 25.83 & 30.11 & 28.99\cr
    \bottomrule
    \end{tabular}
     \caption{\textbf{Super-resolution}. Quantitative comparison (PSNR) on super-resolution (scale $\times$ 4) task in both synthetic and realistic settings. We evaluate the performance of image-level augmentation and YOCO applied to mixture of augmentations (7 in total).   }
     \label{tab:superres}
\end{table}

%% file: tables/std.tex
\begin{table*}[ht!]
  \centering
  \fontsize{8}{3}
  \selectfont
    \begin{tabular}{lc|c|c|c|c|c}
    \toprule
     \Rows{Models}&\multicolumn{2}{c|}{ CIFAR-10}&\multicolumn{2}{c|}{CIFAR-100}&\multicolumn{2}{c}{ImageNet} \cr
    &YOCO &Image-level &YOCO &Image-level &YOCO &Image-level \cr
    \midrule
    PreResNet18& 0.77 & 0.77 & 1.99 & 1.71 & - & - \cr
    Xception& 0.47 & 0.59 & 2.02 & 1.73 & - & - \cr
    DenseNet121& 0.65 & 0.55 & 1.84 &1.72 & - & -    \cr
    ResNext50& 0.55 & 0.67 & 2.02 & 1.77 & - & -    \cr
    WRN-28-10& 0.54 & 0.49 & 1.92 & 1.40& - & -    \cr
    ViT& 0.74 & 0.70 & 2.03 & 1.67& - & -    \cr
    Swin& 0.70 & 0.62 & 1.97 & 1.67& - & -    \cr
    \midrule    
    Resnet50& - & - & - & - & 0.26/0.29 & 0.23/0.33  \cr 
    \midrule
        \vspace{-3mm}
    \end{tabular}
     \caption{Standard deviation of YOCO and Image-level augmentation on every single model. For ImageNet, we report both \textit{best} result (left) and \textit{last} result (right). }
     \label{tab:std}
    \vspace{-3mm}
\end{table*}